\documentclass[11pt]{article}

\usepackage[T1]{fontenc}
\usepackage{lmodern}
\usepackage[margin=1in]{geometry}
\usepackage{cite}
\usepackage{amsmath,amssymb,amsfonts,mathtools}
\usepackage{graphicx}
\usepackage{booktabs}
\usepackage{braket}
\usepackage[hidelinks]{hyperref}
\usepackage{glossaries}
\glsdisablehyper
\newenvironment{keywords}
  {\par\smallskip\noindent\textbf{Keywords: }\ignorespaces}
  {\par\medskip}

\newacronym{ai}{AI}{Artificial Intelligence}
\newacronym{gru}{GRU}{Gated Recurrent Unit}
\newacronym{lstm}{LSTM}{Long Short-Term Memory}
\newacronym{ml}{ML}{Machine Learning}
\newacronym{nisq}{NISQ}{Noisy Intermediate-Scale Quantum}
\newacronym{pru}{PRU}{Prototypical Recurrent Unit}
\newacronym{qnn}{QNN}{Quantum Neural Network}
\newacronym{qml}{QML}{Quantum Machine Learning}
\newacronym{qgru}{QGRU}{Quantum GRU}
\newacronym{qru}{QRU}{Quantum Recurrent Unit}
\newacronym{qpru}{QPRU}{Quantum Prototypical Recurrent Unit}
\newacronym{qlstm}{QLSTM}{Quantum LSTM}
\newacronym{rnn}{RNN}{Recurrent Neural Network}
\newacronym{vqc}{VQC}{Variational Quantum Circuit}
\newacronym{qrnn}{QRNN}{Quantum recurrent Neural Networks}

\newcommand{\FCOut}{\mathbf{FC}_{\text{out}}}
\newcommand{\FCIn}{\mathbf{FC}_{\text{in}}}
\newcommand{\VQCUpdate}{\mathbf{VQC}_{update}}
\newcommand{\VQCOut}{\mathbf{VQC}_{out}}

\begin{document}

\title{A Quantum Variational Approach to Prototypical Recurrent Unit}

\author{%
Mahyar Sadeghi Garjan\textsuperscript{1},
Tommaso Cesari\textsuperscript{1}, and
Michel Barbeau\textsuperscript{2}\\[0.5em]
\small \textsuperscript{1}School of Electrical Engineering and Computer Science,
University of Ottawa, Canada\\
\small \textsuperscript{2}School of Computer Science,
Carleton University, Ottawa, Canada\\[0.25em]
\small ORCID: M.S.G. 0000-0003-0388-8669; T.C. 0000-0001-5010-1094;\\
\small M.B. 0000-0003-3531-4926
}
\date{}

\maketitle

\begin{abstract}
 We introduce a lightweight \gls*{qpru} that requires significantly fewer parameters than both classical recurrent architectures, such as \gls*{lstm} and \gls*{gru}, and quantum variants, including \gls*{qlstm} and \gls*{qgru}. Despite its compact design, the QPRU achieves competitive forecasting performance, matching state-of-the-art baselines while offering important structural and practical advantages, including enhanced scalability and a reduced number of trainable parameters.
\end{abstract}


\section{Introduction}

Learning from sequential data is a fundamental challenge in \gls*{ml}, with applications in finance, natural language processing, healthcare, and signal analysis. Classical models such as \glspl*{rnn} and their improved variants (\gls*{lstm}~\cite{hochreiter1997long}, \gls*{gru}~\cite{cho2014properties}, \gls*{pru}~\cite{long2018prototypical}) address issues like vanishing gradients, long-range dependencies, and model complexity. With the emergence of quantum computing, researchers have explored quantum counterparts, such as \glspl*{qrnn} and \glspl*{qlstm}, which leverage quantum representational capacity and reduced parameterization to model temporal dependencies efficiently, while enabling richer transformations through quantum operations~\cite{Goto2020,Kuebler2021,Raubitzek2023}.

Despite these advances, many quantum recurrent architectures remain complex, resource-intensive, or difficult to train due to circuit depth and measurement overhead. To address this, we propose the \gls*{qpru}, a lightweight, parameter-efficient quantum recurrent unit inspired by the classical \gls*{pru}. The \gls*{qpru} significantly reduces parameter count compared to \gls*{qlstm}, \gls*{qgru}, \gls*{lstm}, \gls*{qru}, and \gls*{pru}, while maintaining forecasting accuracy on time-series benchmarks. 

\paragraph*{Outline} A review of related work is provided in Section~\ref{sec:related-work}.
Section~\ref{sec:background} provides background on \glspl*{vqc} as the main component of our approach.
Section~\ref{sec:method} introduces the proposed \gls*{qpru} model, including its circuit design and update equations.  
Section~\ref{sec:experiments} presents the experimental setup, datasets, and evaluation metrics.  
Section~\ref{sec:results} reports the performance of \gls*{qpru} relative to classical and quantum baselines.  
Section \ref{sec:disscussion} summarizes our main findings and their implications. Finally, Section \ref{sec:conclusion} concludes the work and outlines potential directions for future research.

\section{Related Work}
\label{sec:related-work}
The related work can be broadly divided into classical and quantum recurrent models. 
Since the primary focus of this work is on quantum recurrent architectures, we do not 
review classical models in detail. Readers interested in classical sequence modeling 
and gated recurrent units may refer to the following foundational works:
\cite{pmlr-v28-pascanu13, hochreiter1997long, cho2014properties, Chung2014_GRUvsLSTM, long2018prototypical}.

Despite extensive classical research on time-series forecasting, quantum approaches to sequential modeling remain relatively underexplored. Recent efforts, such as QSegRNN for forecasting real-world time series \cite{Moon2025_QSegRNN}, \glspl*{qrnn} for stock, meteorological and text data \cite{Li2023_QRNN}, continuous-variable quantum RNNs (CV-QRNNs) with faster training \cite{siemaszko2023rapid}, and quantum reservoir computers optimized for classical time series and chaotic dynamics \cite{Kutvonen2020_QRC, ahmed2025optimal}, hint at the potential of quantum sequential modeling.

Quantum Weighted GRU Networks (QWGRUNNs)~\cite{kyriienko2021quantum} replace classical weights with quantum parameters, yielding improved prediction accuracy, but at the cost of substantial quantum resource overhead, which limits their practicality in current \gls*{nisq} devices. Similarly, quantum-weighted \gls*{lstm} variants, including the Mogrifier-QWMELSTM and Parametrized Quantum LSTM (PQ-LSTM)~\cite{chen2022quantum}, achieve good forecasting results but rely on computationally expensive quantum operations.

Other approaches, such as Continuous-Variable Quantum RNNs (CV-QRNNs)~\cite{killoran2019continuous}, offer faster convergence but depend on specialized photonic hardware. Duplication-Free \gls*{qlstm} architectures employ amplitude encoding, which is highly expressive but scales poorly for large datasets. Although these hybrid models improve hardware feasibility, they often remain parameter-heavy or lack comprehensive empirical comparisons against classical baselines.\cite{wang2025limitations}. 

\section{Background}
\label{sec:background}
\glspl*{vqc} are parameterized quantum circuits that have recently emerged as a powerful framework for \gls*{qml} and optimization applications~\cite{Schuld2015, Cerezo2021,Raubitzek2023, Wu2022_Scalable_QNN, Das2023_VQNN_ImageClass}. They consist of trainable angles in single-qubit rotation gates, optimized classically to minimize a cost function, enabling hybrid quantum-classical implementations of \glspl*{qnn} on \gls*{nisq} devices~\cite{Preskill2018}.  

Let $\boldsymbol{x}_t \in \mathbb{R}^m$ denote a classical input at time $t$, and consider a quantum system with $q$ qubits. A quantum feature map $\phi \colon \mathbb{R}^m \rightarrow \mathcal{H}^{2^q}$ embeds the input into a $2^q$-dimensional Hilbert space via $U_\phi(\boldsymbol{x}_t)\ket{0}^{\otimes q} = \ket{\phi(\boldsymbol{x}_t)} = \ket{\psi}$,
with $\ket{0}^{\otimes q}$ the initial state. For continuous features, \textit{angle encoding} maps classical features to qubit rotation angles using gates such as $R_x$ or $R_y$ \cite{Havlicek2019, Schuld2019}.  

A parametrized ansatz $U_W(\boldsymbol{\theta})$, composed of tunable single-qubit rotations and entangling gates, is applied to the encoded state. Single-qubit rotations control amplitudes and phases, while entangling gates such as circular CNOT layers create correlations; repeating layers enhances expressivity~\cite{Kandala2017,Peruzzo2014}. Measurements on an observable $\hat{O}$ yield 
$\hat{f}(\boldsymbol{x}_t, \boldsymbol{\theta}) = \bra{\psi(\boldsymbol{x}_t, \boldsymbol{\theta})}\hat{O}\ket{\psi(\boldsymbol{x}_t, \boldsymbol{\theta})}$, with $\ket{\psi(\boldsymbol{x}_t, \boldsymbol{\theta})} = U_W(\boldsymbol{\theta}) \ket{\phi(\boldsymbol{x}_t)}$. Outputs are obtained in the Pauli-Z basis, and multiple shots account for quantum probabilistic nature.  

Parameters $\boldsymbol{\theta}$ are optimized classically to minimize $C(\hat{f}(\boldsymbol{x}_t, \boldsymbol{\theta}))$, with gradients computed via adjoint differentiation~\cite{Schuld2019grad, Grimsley2019}. Alternating quantum evaluations and classical updates, properly designed encodings and ansatz enable \glspl*{vqc} to capture complex correlations while remaining suitable for \gls*{nisq} devices \cite{Cerezo2021}.
\section{Methodology}
\label{sec:method}
\subsection{Architecture Design}
Our approach incorporates \glspl*{vqc} to model temporal dependencies. Inspired by \gls*{qgru}~\cite{ceschini2024variational} and \gls*{qlstm} \cite{chen2022quantum}, we propose a hybrid architecture in which quantum circuits act as nonlinear transformations within recurrent updates.

\begin{figure}[htbp] 
    \centering
    \includegraphics[width=\linewidth]{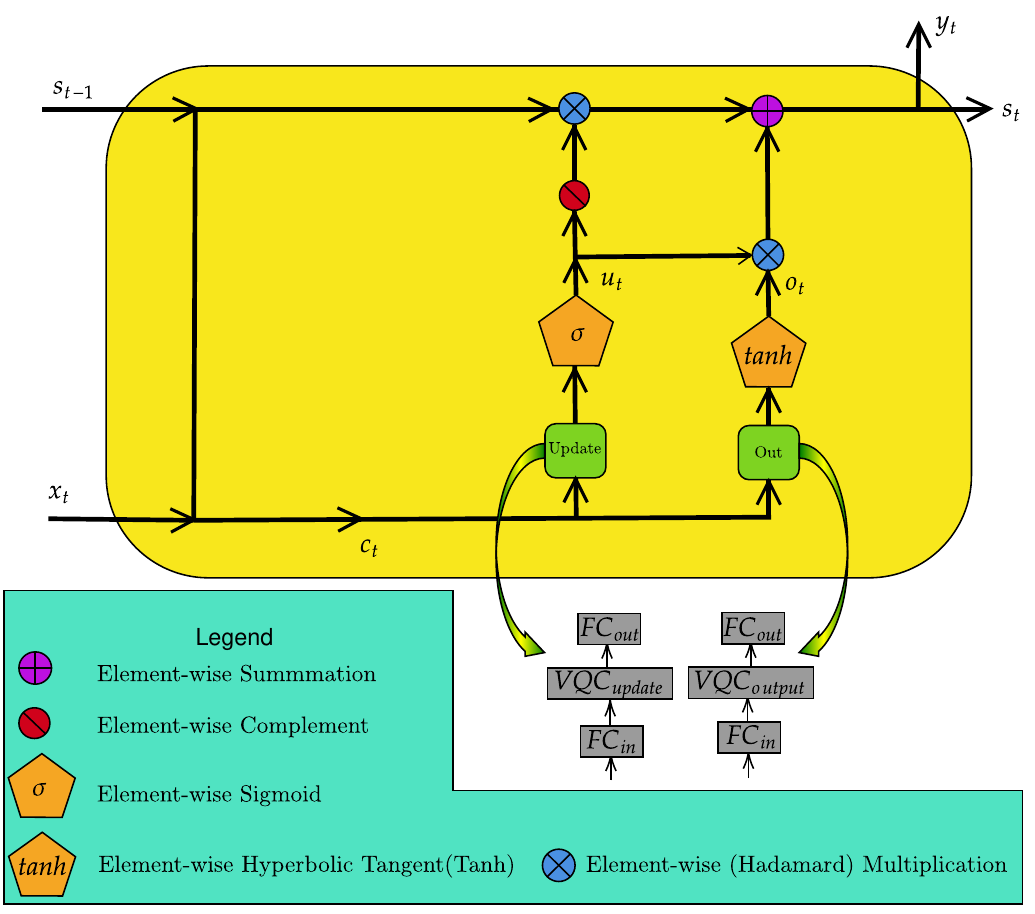}
    \caption{Architecture of the proposed QPRU.}
    \label{fig:qpru_architec}
\end{figure}

\gls*{qpru} is introduced as an optimized alternative to \gls*{qlstm} and \gls*{qgru}, focusing on reducing the number of \glspl*{vqc} exploited while preserving the model's representational capacity. 
The architecture of \gls*{qpru} is shown in Figure~\ref{fig:qpru_architec}.

Let $\boldsymbol{x}_{t} \in \mathbb{R}^m$ be an input column-vector of $m$-dimensional at time $t$  and $\boldsymbol{s}_{t-1} \in \mathbb{R}^k$ be a column-vector of $k$-dimensional representing the hidden state from the previous time step. The concatenation of these two vectors is an $(m+k)$-dimensional column-vector $\boldsymbol{c}_{t} \in \mathbb{R}^{m+k}$ where $\boldsymbol{c}_{t} = [\boldsymbol{s}_{t-1}^\top,\boldsymbol{x}_{t}^\top]^\top$.

Equation~\ref{equ:u_t_computation} shows how to compute the intermediate vector $\boldsymbol{u}_{t}$ using $\boldsymbol{c}_{t}$, through some compositional functions.  
Let $\FCIn{}$ be a classical fully connected layer that takes a $(m+k)$-dimensional vector and maps it to a $q$-dimensional output vector. The resulting output of $\FCIn{}$ is passed to a \gls*{vqc} where quantum learning occurs. Moreover, define $\FCOut{}$ to be a classical fully connected network that takes $q$-dimensional measured qubits as input and maps them to a $k$-dimensional vector. Simply, $\FCOut{}$ maps the $q$-dimensional output of the \gls*{vqc} to a vector of $k$-dimensional using a linear transformation. 
Define $\sigma \colon \mathbb{R}\xrightarrow{}\mathbb{R}$ as a sigmoid function (soft-max or logistic), mapping the input of real value to the output of real value. Applying $\sigma$ element-wise to the output of $\FCOut{}$ results in another vector  $\boldsymbol{u}_{t} \in \mathbb{R}^{k}$ as shown in Equation~\ref{equ:u_t_computation}.
\begin{equation}
    \begin{aligned}
            \boldsymbol{u}_t &= \sigma \Big( \FCOut \Big( \VQCUpdate \Big( \FCIn ((\boldsymbol{c}_t)) \Big) \Big) \Big)
    \end{aligned}
    \label{equ:u_t_computation}
\end{equation}
In Equation~\ref{equ:u_t_computation}, $\boldsymbol{u}_t$ acts as the update gate controlling memory retention.
Similarly, \gls*{qpru} has an output gate, a computing vector $\boldsymbol{o}_{t} \in \mathbb{R}^{k}$ using compositional functions $\FCIn{}$ that takes $\boldsymbol{c}_{t}$ as input and adjusts its dimensionality to the input size of \gls*{vqc}. Using $\FCOut{}$, we map the $q$-dimensional vector to a $k$-dimensional vector. Let $\tanh\colon \mathbb{R}\xrightarrow{}\mathbb{R}$ be a hyperbolic tangent mapping a real-valued input to a real-valued output. The element-wise application of $\tanh$ to the output of $\FCOut{}$ results in $\boldsymbol{o}_{t}$ as shown in Equation~\ref{equ:o_t_computation}. 
\begin{equation}
    \begin{aligned}
            \boldsymbol{o}_t &= \tanh \Big( \FCOut{} \Big( \VQCOut{} \Big( \FCIn{}(\boldsymbol{c}_t) \Big) \Big) \Big)
    \end{aligned}
    \label{equ:o_t_computation}
\end{equation}
Note that to save computational resources, we use $\FCIn{}$ and $\FCOut{}$ in both gates. However, $\VQCUpdate{}$ and $\VQCOut{}$ are different circuits and are trained separately despite their same input-output dimensionalities.  

The hidden state candidate, $\boldsymbol{s}_t$, is computed as in Equation~\ref{equ:s_t_computation}, where $\odot$ is element-wise multiplication.
\begin{equation}
    \begin{aligned}
            \boldsymbol{s}_t &= (1 - \boldsymbol{u}_t) \odot \boldsymbol{s}_{t-1} + \boldsymbol{u}_t \odot \boldsymbol{o}_t
    \end{aligned}
    \label{equ:s_t_computation}
\end{equation}
Equation~\ref{equ:s_t_computation} integrates information from both the past and new input. Note that we do not directly multiply $\boldsymbol{u}_t$ by $\boldsymbol{s}_{t-1}$. Instead, we compute the elemental complement of $\boldsymbol{u}_t$ (denoted by $1- \boldsymbol{u}_t$), then multiply it by  $\boldsymbol{s}_{t-1}$. This strategy leads to a more accurate prediction as shown in~\cite{long2018prototypical}.

A flexible way to compute the prediction $\boldsymbol{y}_t \in \mathbb{R}^{l}$, where $l$ is the dimension of prediction of the model, is to employ a linear transformation using the weight matrix $\boldsymbol{W}_s \in \mathbb{R}^{k\times l}$ and a bias vector $\boldsymbol{b}_s \in \mathbb{R}^{l}$, i.e., $\boldsymbol{y}_t =\boldsymbol{W}_{s}^\top\boldsymbol{s}_{t}+\boldsymbol{b}_{s}$. 

\begin{figure}[htbp] 
    \centering
    \includegraphics[width=\linewidth]{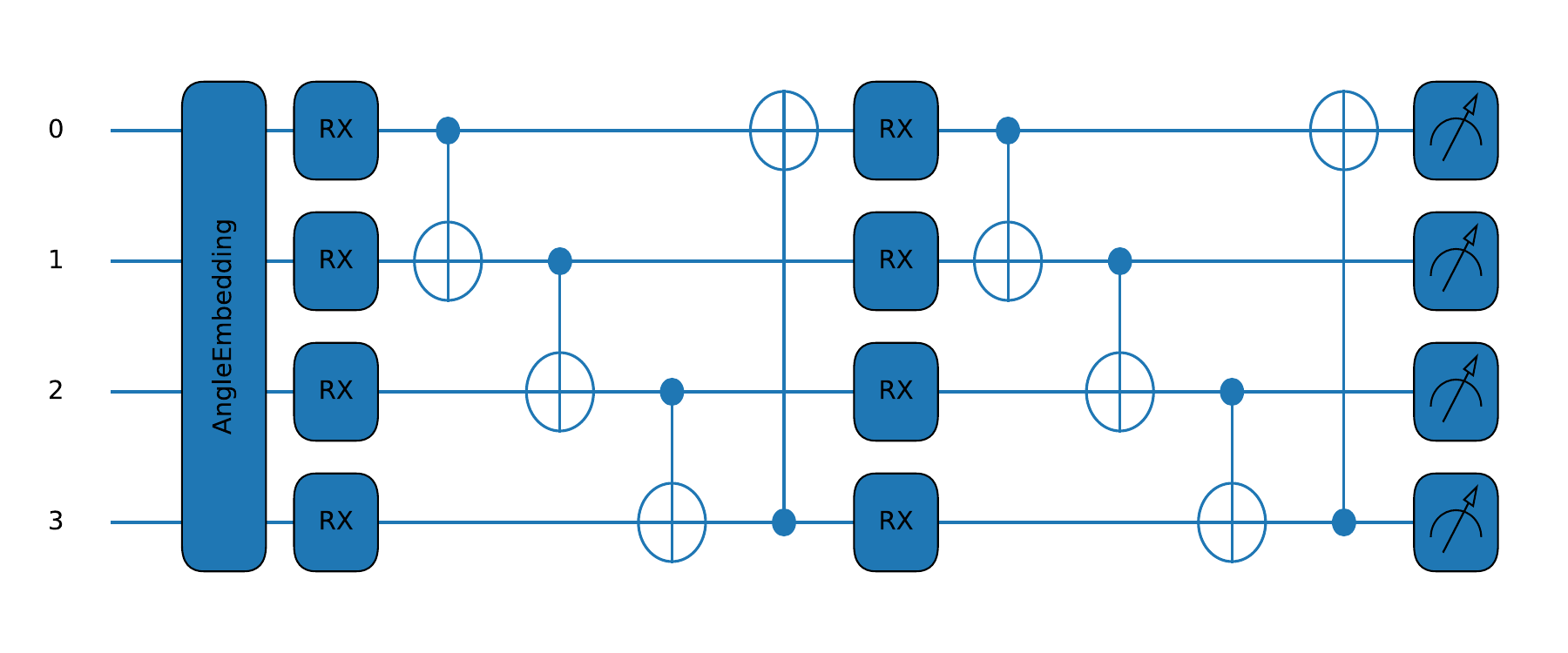}
    \caption{VQC exploited in QPRU.}
    \label{fig:vqc}
\end{figure}

The \gls*{vqc} used in this study is illustrated in Figure~\ref{fig:vqc}. The circuit begins with an angle embedding layer that encodes classical input data into quantum states using rotation gates. This is followed by parameterized single-qubit Pauli-X rotations $R_X(\theta)=e^{-i \theta X / 2}$, which introduce learnable parameters into the model. A layer composed of CNOT gates then creates entanglement between qubits, after which a second parameterized rotation layer further increases circuit expressivity. Finally, all qubits are measured to produce classical outputs.

The \gls*{qpru} cell reduces computational cost through fewer \glspl*{vqc}, parameter sharing, and quantum entanglement, while remaining flexible and scalable, accommodating mismatched input and hidden state dimensions across tasks.

\subsection{Complexity Analysis}
We provide a computational complexity analysis of \gls*{qlstm}, QGRU, and the proposed \gls*{qpru}. Let $d_{\text{hid}}$ denote the hidden state dimension, $q$ the number of qubits, $d_{\text{in}}$ the input feature dimension, and $l$ the number of ansatz layers per \gls*{vqc}. Each \gls*{vqc} contains $q l$ trainable quantum parameters. The classical input layer $\mathrm{FC}_{\text{in}}$ maps a concatenated vector of dimension $d_{\text{conc}} = d_{\text{hid}} + d_{\text{in}}$ to a $q$-dimensional vector, while the classical output layer $\mathrm{FC}_{\text{out}}$ maps from $q$ features to $d_{\text{hid}}$ features.

A \gls*{qlstm} cell contains four \glspl*{vqc} (corresponding to three gates and one memory cell), one $\mathrm{FC}_{\text{in}}$, and one $\mathrm{FC}_{\text{out}}$. Therefore, the total number of parameters is computed as follows:
\begin{align*}
\#\mathrm{params}_{\mathrm{QLSTM}} 
&
= 4 q l 
+ \underbrace{q(d_{\text{hid}} + d_{\text{in}}) + q}_{\mathrm{FC}_{\text{in}}} 
+ \underbrace{q d_{\text{hid}} + d_{\text{hid}}}_{\mathrm{FC}_{\text{out}}} 
\\
&
= q(4 l + 2 d_{\text{hid}} + d_{\text{in}} + 1) + d_{\text{hid}}.
\end{align*}

Similarly, a \gls*{qgru} contains three \glspl*{vqc}, one $\mathrm{FC}_{\text{in}}$, and one $\mathrm{FC}_{\text{out}}$, while the \gls*{qpru} contains two \glspl*{vqc} along with the same classical layers. The total parameters for these architectures can be computed analogously (see Table~\ref{tab:quantum-complexity}). 

From table~\ref{tab:quantum-complexity}, it is evident that \gls*{qpru} contains 50\% fewer quantum parameters than \gls*{qlstm} and 33\% fewer than QGRU, making it the most parameter-efficient among the three architectures. This reduction is particularly significant given the current limitations of quantum simulators and NISQ hardware, where lowering the number of quantum parameters directly improves computational efficiency~\cite{Cerezo2021}.

\begin{table}[!ht]
\centering
\caption{Parameter complexity of QLSTM, QGRU, and the proposed QPRU model.}
\label{tab:quantum-complexity}

\begin{tabular}{lcc}
\toprule
\textbf{Model}  & \textbf{Quantum Params} & \textbf{Total Params} \\
\midrule
QPRU    & $2ql$ & $q(2l + d_{\text{in}} + 2d_{\text{hid}} + 1) + d_{\text{hid}}$ \\
QGRU   & $3ql$ & $q(3l + 2d_{\text{hid}} + d_{\text{in}} + 1) + d_{\text{hid}}$ \\
QLSTM & $4ql$ & $q(4l + 2d_{\text{hid}} + d_{\text{in}} + 1) + d_{\text{hid}}$ \\
\bottomrule
\\
\end{tabular}
\end{table}

\section{Experiments}
\label{sec:experiments}
We conducted two experiments to compare \gls*{qpru} with other recurrent models, namely, \gls*{qlstm}, QGRU, classical \gls*{pru}, \gls*{gru}, and \gls*{lstm}. Each model was trained for 100 epochs, and 20\% of the data was reserved for validation. The models were trained using the Mean Squared Error (MSE) loss. Inspired by~\cite{chen2022quantum,lindsay2023novel}, Root Mean Square Propagation (RMSProp) was chosen as the optimizer. A grid search on the sinusoidal validation set determined the best hyperparameters, fixing the window size to 3, RMSProp smoothing ($\alpha$) to 0.7, learning rate to 0.01, and hidden state size to 5 for all experiments.
Table \ref{tab:hyper-selection-models} lists total trainable parameters for each model, combining classical and quantum components under the grid search configuration.

For each experiment, we report the $\ell_1$, $\ell_2$, and $\ell_\infty$ norms on the validation set, along with their standard deviations.
We used PyTorch (along with PennyLane) as the interface on the Lightning Qubit device with the adjoint differentiation method to accelerate computations.
\begin{table}[!ht]
\centering
\caption{Parameter comparison of classical and quantum recurrent units for hidden state dimension $5$, input feature dimension $1$, $4$ qubits, and $2$ ansatz layers. Classical counterparts are shown in parentheses.}
\label{tab:hyper-selection-models}
\begin{tabular}{lccc}
\toprule
\textbf{Model} & \textbf{Classical}  & \textbf{Quantum} & \textbf{Total} \\
\textbf{} & \textbf{Params}  & \textbf{Params} & \textbf{Params} \\
\midrule
QPRU (PRU)   & $53\,(80)$  & $16\,(0)$ & $69\,(80)$  \\
QGRU (GRU)   & $53\,(120)$ & $24\,(0)$ & $77\,(120)$ \\
QLSTM (LSTM) & $53\,(60)$  & $32\,(0)$ & $85\,(160)$ \\
\bottomrule
\end{tabular}
\end{table}
For Experiment~I, we use a synthetic sinusoidal dataset sampled over the interval $[0, 18\pi]$. 
Overlapping input windows of length 3 were created, with the target being the next value in the sequence. 

Experiment~II uses daily closing prices of AAPL stock from January 1, 2020, to January 1, 2024, obtained from Yahoo Finance. 
Normalization is applied using training-set parameters, scaling each data point to the range $[-1, 1]$. 
The normalized data are segmented into overlapping windows of length 3, with the target being the next closing price.

\section{Results}
\label{sec:results}
\subsection{Results of Experiment I}

Table~\ref{tab:sin-results} summarizes the performance of all models on the sinusoidal benchmark dataset. The metrics reported include the $\ell_1$ norm (MAE), $\ell_2$ norm (MSE), $\ell_\infty$ norm, and their standard deviations, averaged over 10 independent runs for each model. Among all models, \gls*{qpru} achieved the best performance, with the lowest $\ell_1$ norm ($0.012949$), $\ell_2$ norm ($0.000263$), and $\ell_\infty$ norm ($0.0282$) indicating higher accuracy in capturing the underlying sin function. \gls*{qpru} outperforms both classical (\gls*{pru}, \gls*{gru}, \gls*{lstm}) and quantum (\gls*{qgru}, \gls*{qlstm}) models in accuracy and stability, with relatively low standard deviations that demonstrate consistent predictions across runs, highlighting its architectural advantage.

\begin{figure}[htbp] 
    \centering
    \includegraphics[width=\linewidth]{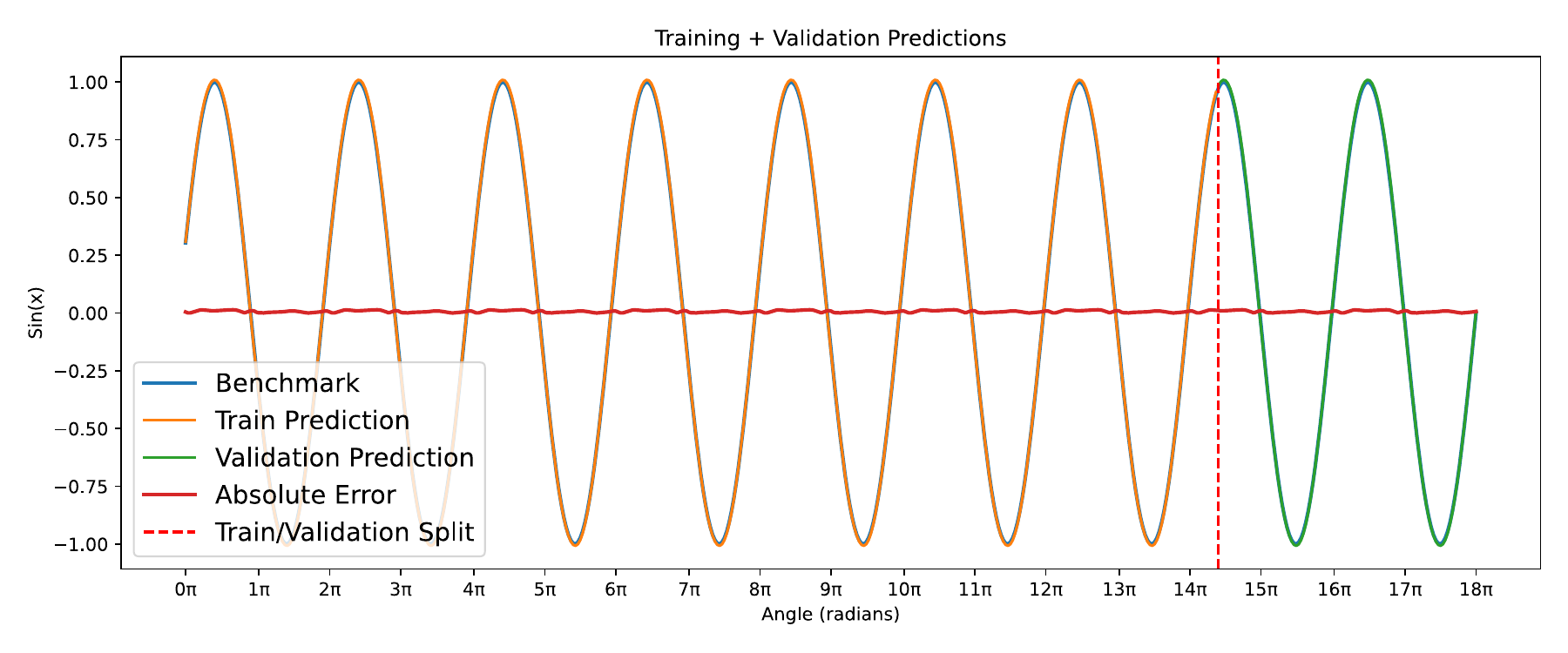}
    \caption{Sinusoidal Function Forecasting using \gls{qpru}. The red solid line represents the absolute error, and the dashed red line represents the training/validation split.}
    \label{fig:training_validation_sin}
\end{figure}

Figure~\ref{fig:training_validation_sin} shows the \gls*{qpru} predictions on the sinusoidal benchmark dataset alongside the original function for both training and validation sets, demonstrating that the model closely tracks the underlying pattern with low absolute error, confirming its ability to learn and generalize periodic sequences.

\begin{table}[!ht]
\centering
\caption{Experiment I results. Norm values are averaged over 10 independent validation runs for each model. Values are expressed in $(mean \pm STD) \times 10^{-n}$ format.}
\label{tab:sin-results}
\begin{tabular}{lccc}
\toprule
\textbf{Model} & $\boldsymbol{\ell_1}(\times 10^{-2}$) & $\boldsymbol{\ell_2}(\times 10^{-4}$) & $\boldsymbol{\ell_\infty}(\times 10^{-2})$ \\
\midrule
QPRU   & $1.2949 \pm 0.4543 $ & $2.63 \pm 1.83 $ & $2.8277\pm 0.9683$ \\
QGRU   & $1.5572 \pm 0.6885 $ & $3.66 \pm 2.59 $ & $3.1338\pm1.3277$ \\
QLSTM  & $1.5772 \pm 0.6093 $ & $4.01 \pm 2.74 $& $3.7471\pm1.3887$ \\
PRU    & $1.4474 \pm 0.5951 $ & $3.52 \pm 3.01 $& $3.2643\pm1.2826$ \\
GRU    & $1.7772 \pm 0.6611 $ & $5.10 \pm 3.48 $ & $3.6358\pm1.2057$\\
LSTM   & $1.5491 \pm 0.4759 $ & $3.36 \pm 1.75$ & $3.0791 \pm 0.8634$\\
\bottomrule
\end{tabular}
\end{table}

\subsection{Results of Experiment II}
\begin{figure}[htbp] 
    \centering
    \includegraphics[width=\linewidth]{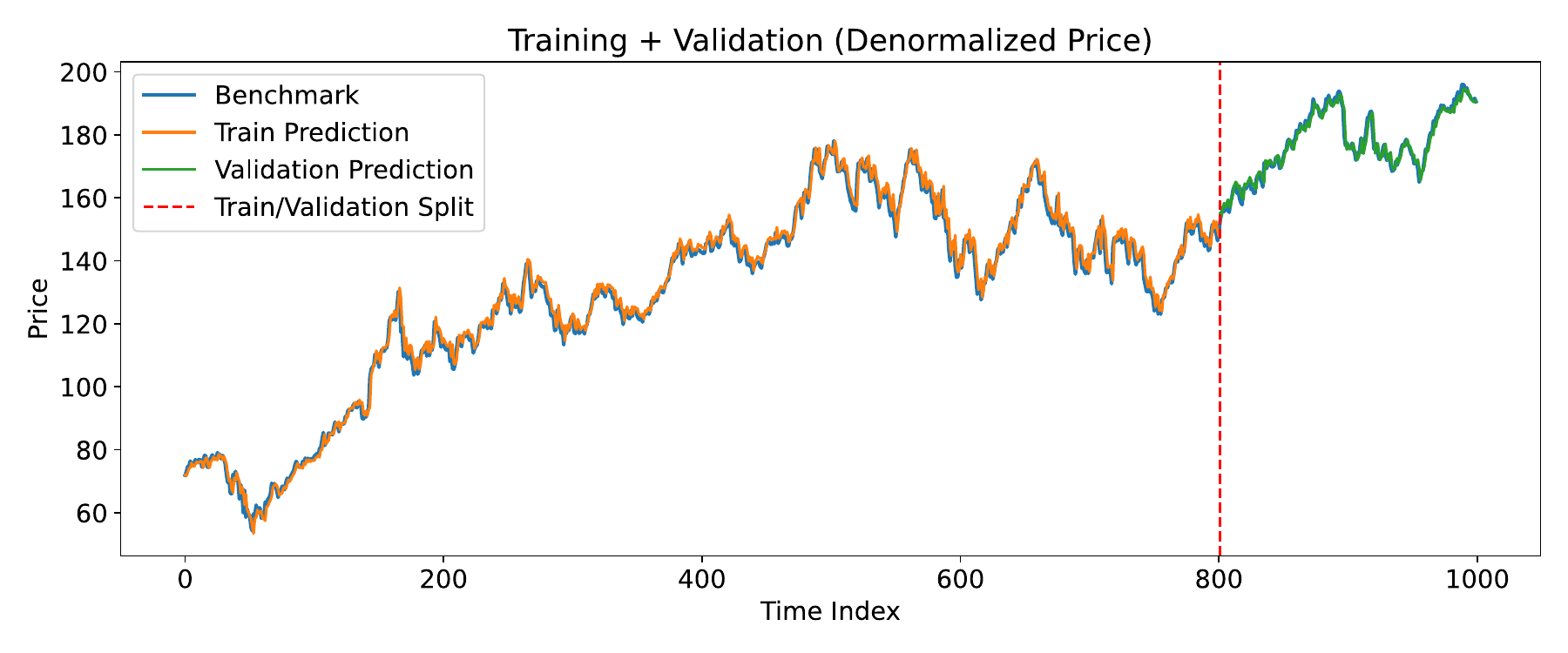}
    \caption{Forecasting of AAPL closing prices. Predictions on training and validation sets are shown, with the dashed line marking the split. Validation MSE and MAE (computed on denormalized data) are $4.4370 \pm 7.920$ and $1.6265 \pm 1.338$, respectively, indicating consistent performance on unseen data.}
    \label{fig:training_validation_stock}
\end{figure}

Figure~\ref{fig:training_validation_stock} illustrates the predictions of the model on the closing prices of the AAPL stock. Similarly to Figure~\ref{fig:training_validation_sin}, the red dashed line indicates the split between the training and validation sets. Data values were normalized to the range $[-1, 1]$ prior to training, ensuring consistent scaling across models.

Table~\ref{tab:stock-results} summarizes the performance metrics for all models in the stock forecasting task. The metrics, including the $\ell_1$ norm, $\ell_2$ norm, $\ell_\infty$ norm were computed on the normalized values and averaged over 10 independent runs. In addition, the standard deviations for the metrics are reported for 10 runs of each model. Among all models, \gls*{qpru} achieved the best performance, exhibiting the lowest $\ell_1$ norm ($0.046131$) and $\ell_2$ norm ($0.003276$), indicating superior accuracy and stability in forecasting the stock price. 

\begin{table}[!ht]
\centering
\caption{Experiment II results. Norm values are averaged over 10 independent validation runs for each model and are computed on data normalized to the range $[-1, 1]$. Values of the table are of the form mean$\pm$ STD.}
\label{tab:stock-results}
\begin{tabular}{lccc}
\toprule
\textbf{Model} & $\boldsymbol{\ell_1}$ & $\boldsymbol{\ell_2}$ & $\boldsymbol{\ell_\infty}$ \\
\midrule
QPRU  & $0.0461\pm0.0151$ & $0.0032\pm0.0017$ & $0.1559\pm0.0203$  \\
QGRU  & $0.1086\pm0.1017$ & $0.0462\pm0.1082$ & $0.3472\pm0.3590$ \\
QLSTM & $0.1198\pm0.0906$ & $0.0552\pm0.0837$ & $0.5473\pm0.5261$ \\
PRU  & $0.0667\pm0.0194$ & $0.0064\pm0.0031$ & $0.1768\pm0.0292$\\
GRU  & $0.0606\pm0.0222$ & $0.0052\pm0.0034$& $0.1573\pm0.0283$\\
LSTM & $0.0553\pm0.0047$ & $0.0065\pm0.0295$  & $0.1816\pm0.1271$\\
\bottomrule
\end{tabular}
\end{table}

\subsection{Hypothesis Testing}
We performed hypothesis tests for the approximation of the $\sin$ and stock curves, comparing our algorithm (denoted here by $k=0$) against each of the five benchmark algorithms described in the previous section ($k = 1,\dots,5)$.
We measure performance as $\ell_p$-distance between each output function and corresponding target curve ($\ell_p$-loss), for $p = 1,2,\infty$, and consider three significance levels: $\alpha = 0.01$, $0.05$, $0.1$.
Denoting by $\mu^{(p)}_j$ the expected $\ell_p$-loss of algorithm $j$, we test, for the three values of $\alpha$ and $p$ introduced above, the null hypothesis
$
H_0^{(p,k)} = \bigl( \mu^{(p)}_0 \ge \mu^{(p)}_k \bigr)
$
against the alternative hypothesis
$
H_1^{(p,k)} = \bigl(\mu^{(p)}_0 < \mu^{(p)}_k\bigr)
$.
In all cases, the tests failed to reject $H_0^{(p,k)}$.
To better understand this outcome, we also considered the reverse comparison:
$
\tilde H_0^{(p,k)} = \bigl( \mu^{(p)}_0 \le \mu^{(p)}_k \bigr)
$
vs.\
$
\tilde H_1^{(p,k)} = \bigl( \mu^{(p)}_0 > \mu^{(p)}_k \bigr),
$
and apply the same testing procedure with the roles of the algorithms swapped.
Again, the tests failed to reject $\tilde H_0^{(p,k)}$ for all $(p,k,\alpha)$.

\begin{figure}[htbp] 
    \centering \includegraphics[width=\linewidth]{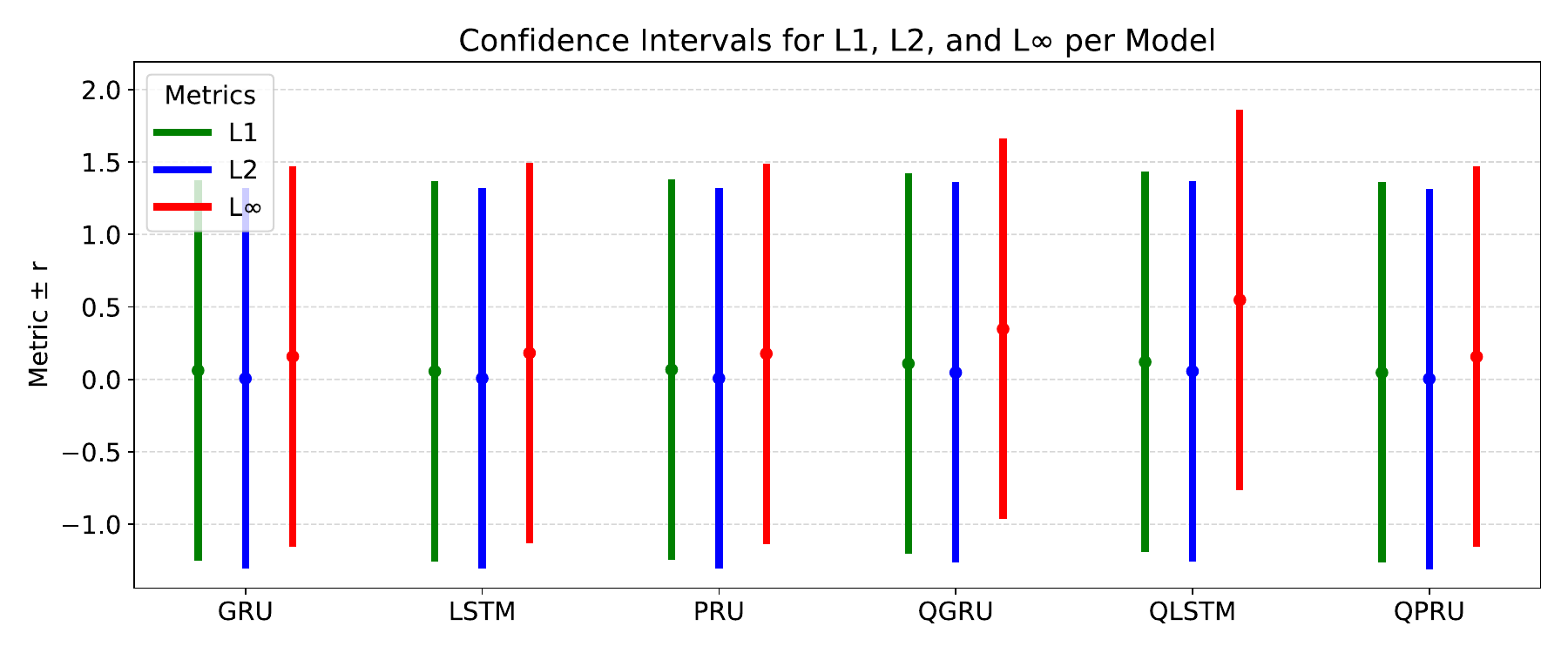}
     \caption{Confidence intervals for stock forecasting for different models. The intervals are computed using 
    $r = 2 \sqrt{\ln(2/\delta)/20}$, with $\delta = 0.0005$, representing the half-width of the interval at a high confidence level. The intervals for all methods overlap, indicating that none of the pairwise differences is statistically significant. Within this experimental setup, our \gls*{qpru} model performs on par with state-of-the-art baselines while retaining the structural and practical advantages discussed earlier.}
    \label{fig:interval}
\end{figure}

This behavior is illustrated in Figure~\ref{fig:interval}.
Note that the confidence intervals around the mean losses of all algorithms overlap, so that none of the pairwise differences is statistically significant at the levels considered.
Within the resolution of this experimental setup, the performance of our algorithm appears statistically indistinguishable from those of the established benchmarks we considered, while retaining the structural and practical advantages discussed in the next section, suggesting that, at least on these tasks and sample sizes, our method performs on par with state-of-the-art baselines.

\section{Discussion}
\label{sec:disscussion}
The results show that \gls*{qpru} achieves strong predictive performance while maintaining a lightweight design. In Experiment~I on the sinusoidal dataset, it obtained the lowest average $\ell_1$, $\ell_2$, and $\ell_\infty$ losses among classical (\gls*{pru}, \gls*{gru}, \gls*{lstm}) and quantum (\gls*{qgru}, \gls*{qlstm}) models, with consistently low standard deviations. It also reduces quantum parameters by approximately 50\% compared to \gls*{qlstm} and 33\% compared to \gls*{qgru} (Table~\ref{tab:quantum-complexity}), an important advantage for NISQ-era devices. In Experiment~II on AAPL stock data, \gls*{qpru} achieved competitive forecasting accuracy, demonstrating its ability to generalize beyond synthetic settings. Overall, the results indicate that a simplified hybrid quantum--classical design can effectively model complex temporal patterns using fewer parameters and shallow circuits.

\section{Conclusion and Future Work}
\label{sec:conclusion}
The \gls*{qpru} design offers several advantages, including reduced quantum resource demands, fewer trainable parameters than its counterparts, stable training behavior, and strong scalability across sequence learning tasks. These features make it particularly suitable for quantum implementations in the \gls*{nisq}-era. 
Future work will explore extending the \gls*{qpru} to multivariate time-series, experimenting with alternative data encoding strategies and ansatz designs to further enhance expressivity, and benchmarking on larger datasets to evaluate performance under more demanding forecasting scenarios. Additionally, testing \gls*{qpru} on actual quantum hardware will provide crucial insight into its practical feasibility and limitations in real-world applications.

\bibliographystyle{unsrt}
\bibliography{references}

\end{document}